\documentclass[runningheads]{llncs}
\usepackage{amssymb}
\usepackage{ragged2e}
\usepackage{amsmath}
\usepackage{stfloats}
\usepackage{booktabs,makecell, multirow}
\usepackage[table]{xcolor}  
\usepackage{hyperref}
\hypersetup{
    colorlinks=true,
    linkcolor=blue,
    citecolor=blue,
    urlcolor=blue
}
\usepackage{marvosym}
\usepackage{ifsym}

\usepackage{bbm}
\usepackage[T1]{fontenc}
\usepackage{graphicx}
\usepackage{color}

\begin{document}

\title{Multi-modal Knowledge Decomposition based Online Distillation for Biomarker Prediction in Breast Cancer Histopathology}
\titlerunning{Multi-modal Knowledge Decomposition based Online Distillation}


\author{Qibin Zhang\inst{1} \and Xinyu Hao \inst{1,2} \and Qiao Chen \inst{2} \and Rui Xu\inst{3} \and Fengyu Cong\inst{1,2} \and Cheng Lu\inst{4}\Letter \and Hongming Xu \inst{1}\Letter}  
\authorrunning{Q. Zhang et al.}
\institute{School of Biomedical Engineering, Faulty of Medicine, Dalian University of Technology, Dalian, China \and Faculty of Information Technology, University of Jyvaskyla, Jyvaskyla, Finland \and School of Software Technology, Dalian University of Technology, Dalian, China \and Department of Radiology, Guangdong Provincial People’s Hospital, Southern Medical University, Guangzhou, China \\
    \email{lucheng@gdph.org.cn; mxu@dlut.edu.cn}}    
    
\maketitle

\begin{abstract}
Immunohistochemical (IHC) biomarker prediction benefits from multi-modal data fusion analysis. However, the simultaneous acquisition of multi-modal data, such as genomic and pathological information, is often challenging due to cost or technical limitations. To address this challenge, we propose an online distillation approach based on Multi-modal Knowledge Decomposition (MKD) to enhance IHC biomarker prediction in haematoxylin and eosin (H\&E) stained histopathology images. This method leverages paired genomic-pathology data during training while enabling inference using either pathology slides alone or both modalities. Two teacher and one student models are developed to extract modality-specific and modality-general features by minimizing the MKD loss. To maintain the internal structural relationships between samples, Similarity-preserving Knowledge Distillation (SKD) is applied. Additionally, Collaborative Learning for Online Distillation (CLOD) facilitates mutual learning between teacher and student models, encouraging diverse and complementary learning dynamics. Experiments on the TCGA-BRCA and in-house QHSU datasets demonstrate that our approach achieves superior performance in IHC biomarker prediction using uni-modal data. Our code is available at \url{https://github.com/qiyuanzz/MICCAI2025_MKD}.
\end{abstract}

\section{Introduction}
Tumor biomarkers, including estrogen receptor (ER), progesterone receptor (PR), and human epidermal
growth factor receptor 2 (HER2), play a critical role in breast cancer diagnosis, therapeutic decision-making, prognostic assessment, and disease monitoring~\cite{wang2024double}. However, determining biomarker status through IHC staining is both time-consuming and costly~\cite{naik2020deep}. With the increasing use of digital histopathology and rapid advancements in deep learning, predicting IHC biomarker status from H\&E-stained whole slide images (WSIs) has emerged as a promising alternative~\cite{naik2020deep,wang2024double}. This method not only offers a more efficient and cost-effective alternative to traditional IHC staining, but also holds significant potential for uncovering complex morphological features intricately linked to biomarker status~\cite{gamble2021determining}.

In recent years, extensive research has focused on leveraging H\&E-stained pathology images to predict tumor biomarker status. Early methods predominantly relied on fully supervised learning, which typically involved training patch-level classifiers based on annotations, with predictions aggregated across the entire WSI to infer patient-level biomarker status~\cite{kather2020pan,naik2020deep}. Since these methods required detailed pixel-level annotations by clinicians, they suffer from a time-intensive and laborious annotation process. The growing adoption of multiple instance learning (MIL) in computational pathology has driven a shift toward weakly supervised learning methods, offering a more efficient and scalable alternative. For instance, Lu et al.~\cite{lu2022slidegraph+} proposed a novel method that utilizes graph convolutional neural networks to extract WSI-level representations for predicting HER2 status, enabling the model to capture the global biological geometric structure of entire slides.  
Wang et al.~\cite{wang2024double} employed CTransPath~\cite{wang2022transformer} as a feature extractor and designed a multi-label learning model capable of simultaneously predicting ER, PR, and HER2 biomarker statuses. This innovative approach highlights the growing potential of combining foundational models with MIL aggregation to enhance biomarker prediction performance. 

Weakly supervised learning models demonstrate great potential in WSI classification, but their performance can often be improved by incorporating supplementary data. Recent studies have shown that joint training with genomic profiles and pathology slides can significantly enhance performance in tasks such as tumor subtyping, survival regression, and tumor grading~\cite{chen2021multimodal,chen2022pan,zhou2023cross,xu2023multimodal,hou2023hybrid}. 
For instance, Chen et al.~\cite{chen2021multimodal} proposed a cross-modal attention mechanism to integrate pathology and genomic features, which enhances the model's ability to capture complex inter-modal relationships. Zhou et al.~\cite{zhou2023cross} developed two parallel encoder-decoder architectures to fuse intra-modal information and generate cross-modal representations, thus improving the model's representational capacity and predictive accuracy. Despite these advancements, clinical application of multi-modal learning is often hindered by the high costs and technical complexities associated with acquiring genomic and pathology data simultaneously.  

In order to mitigate the need for expensive genomic data collection, we propose an approach that leverages multi-modal data during training while enabling inference using only pathology slides. To handle missing modalities during testing, we employ knowledge distillation (KD)~\cite{hinton2015distillingknowledgeneuralnetwork}, a widely-used technique for transferring knowledge from a multi-modal teacher to a uni-modal student~\cite{xing2022discrepancy}. However, several challenges remain when applying KD to our scenario: (1) Teacher models are often selected based on subjective experience rather than objective metrics, leading to suboptimal guidance and inconsistent distillation outcomes. (2) According to the \textit{Modality Focusing Hypothesis} (MFH)~\cite{xue2022modality}, the effectiveness of cross-modal KD depends on the retention of modality-general decisive features in the teacher model. A higher proportion of such features enhances student model performance, yet effective strategies to increase this proportion remain underexplored and challenging.

To address the aforementioned challenges, we propose an online KD approach based on multi-modal knowledge decomposition (MKD), designed to amplify the representation of modality-general decisive features while concurrently optimizing the efficacy of both student and teacher models. Our main contributions are: (1) We employ the MKD to enhance the transferability of modality-general decisive features from the teacher model. (2) We propose a robust and efficient online KD model that ensures the teacher and student models appropriately capture the relative relationships among samples, while fostering greater diversity and complementarity in dynamic learning. (3) Extensive experiments show that both the teacher and student models achieve state-of-the-art (SOTA) performance on both the public TCGA-BRCA and in-house datasets in biomarker prediction.

\begin{figure}[!htbp]
\centering
\resizebox{0.9\textwidth}{!}{\includegraphics{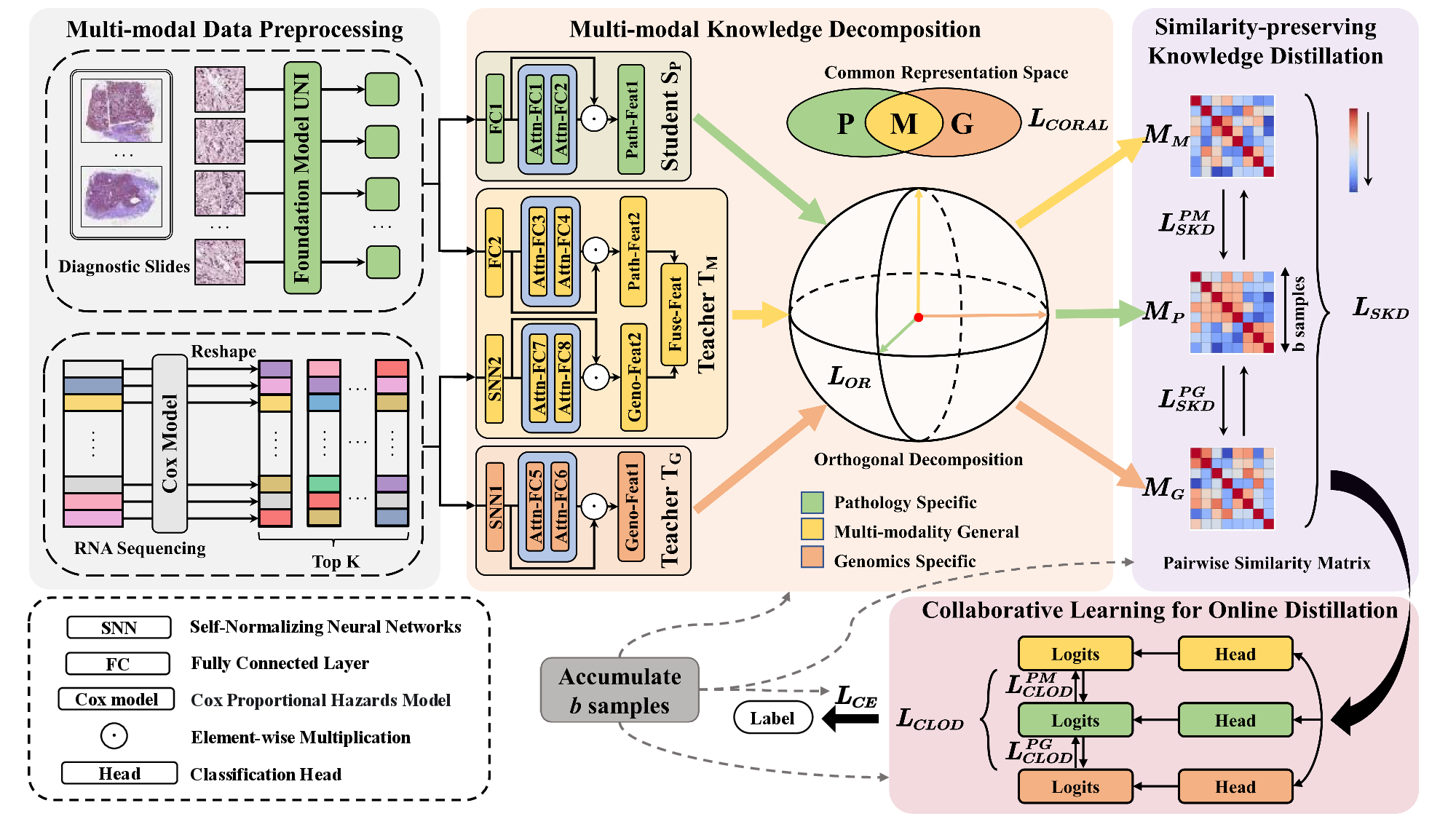}}
\caption{Overview of our approach. Note that our approach includes four components: multi-modal data preprocessing (MDP), MKD, SKD, and CLOD.}
\label{framework}
\end{figure}

\section{Methods}
Fig.~\ref{framework} shows an overview of the proposed approach, including four components: MDP, MKD, SKD, and CLOD. The process starts by extracting genomic and pathomic features, then decomposing multi-modal knowledge into pathology-specific, modality-general, and genomics-specific features. SKD enables the pathology student model to learn sample relationships, while CLOD fosters mutual learning between teacher and student models. During training, gradients accumulate over multiple samples before updating model parameters, allowing SKD to capture feature relationships. Details are provided below.

\subsection{Multi-modal Data Preprocessing (MDP)}
Given a WSI, it is divided into multiple tiles, with tissue tiles selected using the CLAM toolbox~\cite{lu2021data}. Feature embedding is then performed on tissue tiles using the UNI foundation model~\cite{chen2024towards}. Let $P\in \mathbb{R}^{n_p\times d}$ denote the feature embedding for a WSI, where $n_p$ is the number of tissue tiles, and $d$ denotes the feature dimension. We hypothesize that genes associated with overall survival (OS) of patients are also associated with IHC biomarker status. Thus, the Cox proportional hazards model~\cite{cox1972regression} is employed to identify the top $K$ genes most relevant to patients' OS from genomic profiles, with the genomic features reshaped into a representation $G \in \mathbb{R}^{n_g \times d}$ based on their rankings, where $n_g=\left\lfloor {K/d} \right\rfloor $.

\subsection{Multi-modal Knowledge Decomposition (MKD)}
To thoroughly decompose and integrate knowledge, we develop two teacher and one student models, each tailored to capture distinct aspects of genomic profiles and pathology slides. These aggregators focus on collaboration, uniqueness, and general representation, facilitating a comprehensive understanding of multi-modal data. As observed in Fig.~\ref{framework}, pathology features $P$ are first compressed using a fully connected layer, followed by further compression using the Attention-based MIL (ABMIL)~\cite{ilse2018attention} in the student model $S_P$, as expressed as:
\begin{equation}
z_p=\sum\limits_i^{n_p}{a_iP_i}, \quad a_i=\frac{\exp \left\{ W^T\left( \tan\text{h}\left( VP_{i}^{T} \right) \odot \text{sigmoid}\left( UP_{i}^{T} \right) \right) \right\}}{\sum_{j=1}^{n_p}{\exp \left\{ W^T\left( \tan\text{h}\left( VP_{j}^{T} \right) \odot \text{sigmoid}\left( UP_{j}^{T} \right) \right) \right\}}},
\end{equation}
where $V, U\in \mathbbm{R}^{n_p\times d}$ and $W\in \mathbbm{R}^{d\times 1} $ are learnable linear projection matrices, and $\odot$ is an element-wise multiplication. Similarly, genomic features $G$ undergo a two-step process in the teacher model $T_G$: they are first compressed via a Self-Normalizing Network (SNN)~\cite{klambauer2017self}, followed by refinement with ABMIL. Meanwhile, we build a teacher model $T_M$, which fuse the global representations of two modalities learned by ABMIL using the Kronecker product~\cite{van2000ubiquitous}. By processing through the three distinct aggregators including pathology-specific, modality-general, and genomic-specific features, the multi-modal knowledge are systematically decomposed, enabling the extraction of an integrated and meaningful knowledge representation.

To advance knowledge distillation and enhance the model's generalization ability, we perform domain alignment on the decomposed knowledge. Specifically, we minimize the CORAL loss~\cite{sun2016deep} between the decomposed knowledge, which is expressed as:
\begin{equation}
L_{CORAL}=\frac{1}{4d^2}\left( \lVert C_{P}^{b}-\left. C_{G}^{b} \rVert _{F}^{2}+\lVert C_{P}^{b}-\left. C_{M}^{b} \rVert _{F}^{2}+\lVert C_{G}^{b}-\left. C_{M}^{b} \rVert _{F}^{2} \right. \right. \right. \right), 
\end{equation}
where $\lVert \cdot \rVert _{F}$ denotes the Frobenius norm, and $C_{j}^{b}\in \mathbb{R}^{d\times d}$ is the covariance matrix for $b$ cumulative samples, where $j\in \{P,G,M\}$. The $L_{CORAL}$ loss reduces covariance differences across modalities, aligning features into a unified representation for compatible decomposed knowledge. To transfer decisive modality-general features to the student model $S_P$, we introduce a pairwise orthogonality constraint, promoting feature independence and ensuring each captures distinct, complementary information from different modalities. To enforce this, we introduce an orthogonal loss function as follows:
\begin{equation}
L_{OR}=\left| \left< z_p,z_g \right> \right|+\left| \left< z_p,z_m \right> \right|+\left| \left< z_g,z_m \right> \right|,
\end{equation}
where $\left< \cdot \right>$ denote the dot product, and $\left| \cdot \right|$ is the absolute operation used to enforce pairwise orthogonality between inputs. $z_p,z_g,z_m\in \mathbb{R}^{1\times d}$ represent the features specific to pathology, genomics and multi-modal general features, respectively. This orthogonal loss reduces redundancy and promotes more robust feature representations across modalities. Consequently, our MKD loss is calculated as:
\begin{equation}
    L_{MKD}=L_{CORAL}+\alpha L_{OR},
    \label{alpha}
\end{equation}
where $\alpha$ is a hyperparameter that serves as a weighting parameter.

\subsection{Similarity-preserving Knowledge Distillation (SKD)}
To capture sample relationships, we introduce SKD~\cite{tung2019similarity} to guide the training of the student model $S_P$, ensuring input pairs with similar (or dissimilar) activations in the teacher network also produce similar (or dissimilar) activations in the student network. Our SKD loss is calculated as:
\begin{equation}
L_{SKD}=L_{SKD}^{PM}+L_{SKD}^{PG},
\end{equation}
where the $L_{SKD}^{PM}$ loss for a single pair of student and teacher models is computed as:
\begin{equation}
L_{SKD}^{PM}=\frac{1}{b^2}\lVert \frac{Z_{P}^{b}\left( Z_{P}^{b} \right) ^T}{\lVert Z_{P}^{b}\left( Z_{P}^{b} \right) ^T \rVert _2}-\frac{Z_{M}^{b}\left( Z_{M}^{b} \right) ^T}{\lVert Z_{M}^{b}\left( Z_{M}^{b} \right) ^T \rVert _2} \rVert _{F}^{2},
\label{loss_sp}
\end{equation}
where $\lVert \cdot \rVert_2$ denotes row-wise L2 normalization, and $Z_{p}^{b}$, $Z_{M}^{b}$ denote the pathological and multi-modal feature matrices for $b$ concatenated samples. The loss $L_{SKD}^{PG}$ is computed similarly as in Eq.(\ref{loss_sp}). 
Notably, our comprehensive SKD loss effectively preserves the consistency of the activation similarity matrices $M_M$, $M_P$, $M_G$, enabling effective knowledge transfer within multi-modal feature space while maintaining their internal sample-specific structures.

\subsection{Collaborative Learning for Online Distillation (CLOD)}
To foster collaborative learning between teacher and student models, we adopt an online learning framework~\cite{zhang2018deep,zhu2018knowledge}, which consolidates training into a single stage by treating all networks as peers. This framework facilitates symmetrical knowledge sharing, allowing each network learn equally from others without being overly reliant on a predefined teacher model. Our CLOD loss is defined as:
\begin{equation}
    L_{CLOD}=KL\left( p_P||p_M \right) +KL\left( p_M||p_P \right) +KL\left( p_P||p_G \right) +KL\left( p_G||p_P \right),
    \label{kl}
\end{equation}
where $p_P, p_M, p_G$ denote the probability distributions predicted by the pathology-specific, modality-general, and genomic-specific classification heads, respectively. The Kullback-Leibler (KL) divergence terms in Eq.(\ref{kl}) measure distribution discrepancies, promoting alignment and mutual knowledge sharing across networks. This collaborative setup fosters diverse learning dynamics and enables bidirectional knowledge exchange, improving overall performance. Therefore, the overall loss $L$ of our online KD model is formulated as:
\begin{equation}
    L=L_{CE}+L_{MKD}+L_{SKD}+L_{CLOD},
\end{equation}
where $L_{CE}$ represents the overall cross-entropy loss calculated for two teacher models and one student model.

\section{Experiments and Results}
\subsection{Datasets and Implementations}

\begin{figure}[!htbp]
\centering
\resizebox{0.9\textwidth}{!}{\includegraphics{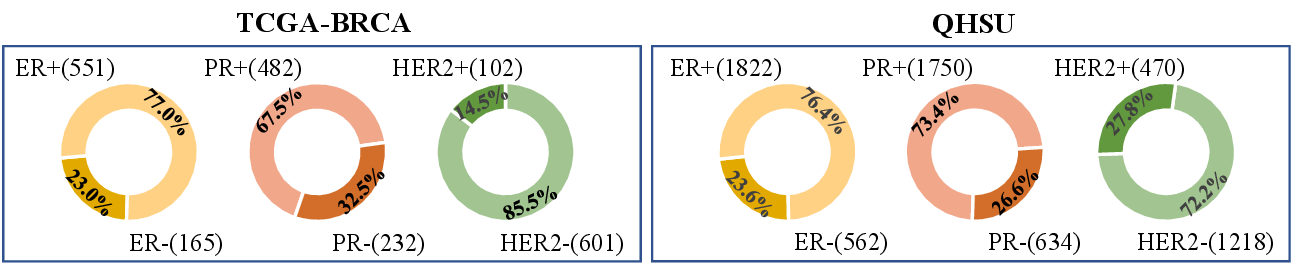}}
\caption{Summary of public TCGA-BRCR and in-house QHSU cohorts.}
\label{dataset}
\end{figure}

\textbf{TCGA-BRCA:} The TCGA-BRCA dataset provides a multi-omics resource, with cases having missing or low-quality genomic profiles or pathology slides excluded. For patients with multiple slides, one diagnostic slide was randomly selected. Genomic profiles were represented using log-transformed, Z-score normalized RNA-Seq expression values. In Fig.~\ref{dataset}, the left three ring charts illustrate the distribution of patients across the ER, PR, and HER2 labels in this dataset, which are used for both internal training and testing cohorts.

\textbf{QHSU:} QHSU dataset includes 2384 H\&E-stained WSIs, with each WSI corresponding to an individual breast cancer patient. IHC biomarker information for these patients was obtained from diagnostic records assessed by skilled pathologists. In Fig.~\ref{dataset}, the right three ring charts illustrate the distribution of patients across the ER, PR, and HER2 labels. Since this in-house dataset contains only pathology slides, it is used as an external test set. 

\textbf{Evaluation \& Implementation:} We performed a 5-fold cross-validation on the TCGA-BRCA internal cohort and report the average test performance across all folds. The five trained models were then tested on the external QHSU cohort, with average results reported. Our model was implemented in Python using the PyTorch library, and trained on a workstation equipped with an NVIDIA GeForce RTX 4090 GPU. We used the AdamW optimizer with a learning rate of 2e-4, a weight decay of 1e-5, and a temperature of 4. The hyperparameter $\alpha$ in Eq.(\ref{alpha}) was fine-tuned and set to 1/6. Model parameters were updated after accumulating gradients from 16 samples.

\subsection{Experimental Results}
\textbf{Internal Comparison.} The proposed model was compared against seven MIL methods~\cite{ilse2018attention,lu2021data,shao2021transmil,zhang2022dtfd,li2024dynamic,tang2024feature,wang2024double}, two KD approaches~\cite{wu2024genomics,xing2024comprehensive}, and three multi-modal learning models~\cite{chen2021multimodal,chen2022pan,zhou2023cross} to evaluate its effectiveness and flexibility. As presented in Table~\ref{TCGA}, the internal comparisons on the TCGA-BRCA cohort reveal that our model, when using only pathology slides, achieves the best overall performance compared to SOTA methods, with a 2\% improvement in AUC values and notable improvements in other metrics. This highlights the capability of the teacher model to effectively transfer critical and generalizable features to the student model, thereby enhancing its performance. In multi-modal testing, all models perform better than using pathology slides alone, particularly for HER2 prediction. Notably, our model consistently ranks first or second across various metrics, indicating that joint training of teacher and student models fosters mutual learning without compromising individual performance. In addition, our approach eliminates the need to explicitly validate the teacher’s guidance, significantly reducing computational costs.

\begin{table}[!htbp]
    \centering
    \caption{Comparisons with SOTA methods on the TCGA-BRCA dataset. The bold and underlined fonts highlight the best and the second-best results, respectively.}
    \label{TCGA}
    \arrayrulecolor{gray}{
    \scalebox{0.9}{\begin{tabular}{ccccccccccc}
    \hline
        {Modality} &{Models}& \multicolumn{3}{c}{ER(\%)} & \multicolumn{3}{c}{PR(\%)} & \multicolumn{3}{c}{HER2(\%)} \\ \cline{3-11}
        ~ & ~ & AUC & ACC &F1 & AUC & ACC &F1 & AUC & ACC &F1\\ \hline 
        Patho.& ABMIL~\cite{ilse2018attention} & 88.91 & 84.85 & 89.84 & 86.10 & 80.55 & 85.08 & 72.13 & 77.30 & 39.36 \\ 
        & CLAM~\cite{lu2021data} & 89.20 & 85.69 & 90.68 & 85.33 & \underline{82.22} & \underline{86.86} & 70.51 & 80.56 & 37.13 \\ 
        & TransMIL~\cite{shao2021transmil} & 87.77 & \underline{86.94} & \underline{91.82} & 81.37 & 77.50 & 83.04 & 66.23 & 75.60 & 33.45 \\ 
        & DTFD~\cite{zhang2022dtfd} & 89.89 & 86.66 & 91.37 & 86.50 & 82.36 & 86.76 & 70.72 & 75.88 & 40.18 \\ 
        & WIKG~\cite{li2024dynamic} & 88.05 & 84.44 & 90.10 & 85.54 & 78.88 & 70.37 & 68.50 & 74.89 & 38.30 \\ 
        & RRT~\cite{tang2024feature} & 89.35 & 83.33 & 88.28 & \underline{86.61} & 80.55 & 85.31 & 68.08 & 77.44 & 28.13 \\ 
        & DAMLN~\cite{wang2024double} & 89.58 & 85.97 & 91.39 & 86.38 & 81.80 &86.54 &69.92 & \textbf{84.53} & 22.37\\
        & GEE~\cite{wu2024genomics} & \underline{90.01} & 84.30 & 89.37 & 86.24 & 81.52 & 85.81 & 70.06 & 74.60 & \underline{41.32}\\ 
        & TDC~\cite{xing2024comprehensive} & 89.35 & 83.33 & 88.28 & 84.75 & 73.05 & 77.01 & \underline{72.86} & 77.87 & \textbf{42.87} \\ 
        & Ours & \textbf{93.31} & \textbf{88.47} & \textbf{92.58} & \textbf{88.65} & \textbf{83.75} & \textbf{88.14} & \textbf{74.56} & \underline{81.56} & 39.10 \\
        \cline{1-11}
        Multi. & MCAT~\cite{chen2021multimodal} & \underline{94.64} & \textbf{90.41} & \textbf{93.68} & 90.37 & \underline{85.28} & 88.75 & \underline{93.10} & 84.96 & 65.06 \\ 
        & Porpoise~\cite{chen2022pan} & 92.64 & 89.86 & 93.28 & \textbf{91.79} & \textbf{85.97} & \textbf{89.57} & 92.80 & \underline{89.65} & \underline{68.94} \\ 
        & CMTA~\cite{zhou2023cross} & 93.91 & 89.44 & 93.15 & \underline{91.51} & 83.63 & 87.26 & 92.03 & 87.94 & 66.55 \\ 
        & Ours & \textbf{95.81} & \underline{90.24} & \underline{93.67} & 91.39 & \underline{85.28} & \underline{89.26} & \textbf{95.76} & \textbf{92.48} & \textbf{76.88} \\ 
        \hline
    \end{tabular}
    }}
\end{table}

\begin{table}[!htbp]
    \centering
    \caption{Comparisons with SOTA methods on the QHSU dataset.}
    \label{QHSU}
    \arrayrulecolor{gray}{
    \scalebox{0.9}{\begin{tabular}{ccccccccccc}
    \hline
        {Modality} &{Models}& \multicolumn{3}{c}{ER(\%)} & \multicolumn{3}{c}{PR(\%)} & \multicolumn{3}{c}{HER2(\%)} \\ \cline{3-11}
        ~ & ~ & AUC & ACC &F1 & AUC & ACC &F1 & AUC & ACC &F1\\ \hline 
        Patho.& ABMIL~\cite{ilse2018attention} & 87.27 & 83.09 & 89.55 & 82.71 & 64.63 & 69.67 & 72.41 & 56.92 & 50.76 \\ 
        & CLAM~\cite{lu2021data} & 86.83 & 82.71 & 89.16 & 80.93 & 64.01 & 68.73 & \underline{73.67} & \underline{60.91} & 49.43 \\ 
        & TransMIL~\cite{shao2021transmil} & 84.74 & 81.24 & 88.21 & 78.33 & \underline{71.42} & \underline{78.48} & 68.95 & 58.83 & 42.49 \\ 
        & DTFD~\cite{zhang2022dtfd} & 87.50 & 82.69 & 89.41 & 81.64 & 61.85 & 65.73 & 71.90 & 56.93 & 50.70 \\ 
        & WIKG~\cite{li2024dynamic} & 82.79 & 84.44 & \underline{90.10} & 79.04 & 58.48 & 60.23 & 70.38 & 52.50 & 49.26 \\ 
        & RRT~\cite{tang2024feature} & 85.31 & 81.44 & 88.65 & 79.09 & 67.13 & 72.98 & 71.81 & 57.17 & 49.78 \\ 
        & DAMLN~\cite{wang2024double} & 86.02 & 82.45 & 88.58 & 81.18 & 70.34 & 76.57 &73.40 & \textbf{69.44} & 42.71\\
        \cline{2-11}
        & GEE~\cite{wu2024genomics} & \underline{88.31} & \underline{83.30} & 87.95 & 82.94 & 70.40 & 76.17 & 71.09 & 49.53 & 48.28 \\ 
        & TDC~\cite{xing2024comprehensive} & 87.77 & 78.48 & 84.31 & \underline{83.40} & 59.19 & 62.72 & 68.88 & 59.38 & \underline{51.29}\\ 
        & Ours & \textbf{89.00} & \textbf{84.97} & \textbf{90.45} & \textbf{84.36} & \textbf{74.01} & \textbf{79.95} & \textbf{74.12} & 57.79 & \textbf{52.74} \\
    \hline
    \end{tabular}
    }}
\end{table}

\textbf{External Comparison.} Table~\ref{QHSU} shows the external comparison results on the QHSU dataset. The results reveal that models trained with knowledge distillation generally outperform MIL models trained solely on pathology slides in ER and PR predictions, indicating that knowledge distillation enables student models to learn more robust features. Notably, our model outperform all comparative methods on the external test set, demonstrating the effectiveness of our multi-modal knowledge decomposition in extracting general decisive features.

\begin{table}[!htbp]
    \centering
    \caption{Ablation study of MKD, SKD, CLOD modules on the TCGA-BRCA dataset.}
    \label{Ablation}
    \arrayrulecolor{gray}{
    \scalebox{0.9}{\begin{tabular}{ccccccccc}
    \hline
        {Modality} &\multicolumn{2}{c}{Modules}& \multicolumn{2}{c}{ER(\%)} & \multicolumn{2}{c}{PR(\%)} & \multicolumn{2}{c}{HER2(\%)} \\ \cline{2-9}
        ~ & MKD & SKD+CLOD &AUC & ACC & AUC & ACC  & AUC & ACC \\ \hline 
        Patho.& ~ & ~ & 88.91 & 84.85 & 86.10 & 80.55 & 72.13 & 77.30\\ 
        & \checkmark & ~ & \underline{93.00} & \underline{86.94} & \underline{87.70} & \underline{84.02} & \underline{72.19} & \underline{82.26} \\
        & ~ & \checkmark & 91.80 & 85.97 & 87.58 & 83.19 & 67.00 & 80.00 \\
        & \checkmark & \checkmark & \textbf{93.31} & \textbf{88.47} & \textbf{88.65} & \textbf{83.75} & \textbf{74.56} & \textbf{81.56 }\\
        \cline{1-9}
        Multi.& ~ & ~ & 94.45 & \textbf{90.97} & 90.14 & 84.16 & 93.57 & 90.78 \\
        & \checkmark & ~ & 91.96 & 89.58 & 88.90 & \underline{84.86} & 92.48 & \textbf{94.18} \\
        & ~ & \checkmark & \underline{95.75} & \underline{90.55} & \underline{91.05} & 84.02 & \underline{95.57} & \underline{93.61} \\
        & \checkmark & \checkmark & \textbf{95.81} & 90.24 & \textbf{91.39} & \textbf{85.28} & \textbf{95.76} & 92.48 \\
    \hline
    \end{tabular}
    }}
\end{table}

\textbf{Ablation Study.} We conducted ablation experiments on the TCGA-BRCA dataset to assess the contributions of the MKD, SKD and CLOD modules to overall model performance. Table~\ref{Ablation} presents evaluation results. It is observed that the MKD module greatly enhances the performance when utilizing pathology slides alone. In contrast, the SKD and CLOD modules, as knowledge distillation components, significantly improve the performance of multi-modal models. Overall, the inclusion of all three modules greatly facilitates mutual learning between the teacher and student models, leading to consistently enhanced performance across different evaluation metrics.

\section{Conclusion}
In this paper, we propose a multi-modal knowledge decomposition based online distillation method for predicting multiple IHC biomarkers from H\&E-stained breast cancer WSIs. Our key innovation lies in the MKD module, which efficiently decomposes input features into pathology-specific, modality-general, and genomics-specific components, facilitating the transfer of both generalizable and decisive knowledge. Additionally, the SKD module enhances knowledge transfer across samples by preserving their internal structures, while the CLOD module fosters mutual learning and knowledge sharing between the teacher and student models. Experiments conducted on two datasets demonstrate that our method achieves superior performance in both uni-modal data testing. Our approach is highly flexible, as its inference supports pathology slides, genomics profiles, or both, depending on available modalities.

\subsubsection{\ackname}
This work was supported in part by Liaoning Province Science and Technology Joint Program (2024-MSLH-065), and the Fundamental Research Funds for Central Universities (DUT25Z2514, DUT24YG201).

\subsubsection{\discintname}
The authors have no competing interests to declare that are relevant to the content of this article.

\bibliographystyle{splncs04}
\bibliography{ref}
\end{document}